\documentclass[letterpaper]{article} 
\usepackage[preprint]{aaai2027} 
\usepackage[hyphens]{url} 
\usepackage{graphicx} 
\usepackage{natbib} 
\usepackage{caption} 
\usepackage{tikz}
\usepackage{amsmath,amssymb}
\usepackage{booktabs}
\usepackage{multirow}
\usepackage{tabularx}
\usetikzlibrary{arrows.meta,positioning,calc}
\title{Beyond KV Reconstruction:\\
Functional Reconstruction for MLA Draft Models in Speculative Decoding}
\author{
Weiye Shi\textsuperscript{1},
Fanxu Meng\textsuperscript{1},
Muhan Zhang\textsuperscript{1,\textdagger}
}
\affiliations{
\textsuperscript{1}Institute for Artificial Intelligence, Peking University\\
\textsuperscript{\textdagger}Corresponding author
}

\begin{document}
\maketitle

\begin{abstract}
Multi-head latent attention (MLA) is an increasingly important architecture for long-context LLM inference, as it replaces the growing key-value (KV) cache with compact latent states and thereby reduces memory traffic during decoding. Yet most capable open checkpoints were pretrained with multi-head or grouped-query attention (MHA/GQA), making conversion to MLA essential for realizing these cache-efficiency benefits without costly retraining from scratch. Meanwhile, speculative decoding has emerged as a complementary approach to accelerate autoregressive generation. Its practical speedup, however, depends critically on agreement between draft proposals and target verification: stochastic sampling is governed by distributional overlap, whereas greedy decoding requires top-ranked-token agreement. We find that direct MHA/GQA-to-MLA conversion can severely degrade this agreement. Low-rank factorization and RoPE handling introduce attention-function errors that may be tolerable in standalone generation, but cause sharply lower token acceptance in speculative decoding. We therefore formulate MLA draft construction as a functional reconstruction problem rather than a cache-compression problem. Our end-to-end (E2E) reconstruction directly optimizes each converted MLA attention module to reproduce the post-output-projection response of its original MHA/GQA counterpart on calibration hidden states. It is a converter-agnostic post-conversion procedure that preserves the converted cache structure and uses no verifier-model logits or supervision. We evaluate 192 model--converter--backend--method--task configurations spanning four Llama/Qwen draft--target pairs, TransMLA and MHA2MLA, HF and vLLM backends, and four 200-prompt tasks. Using a 0.5-percentage-point reporting tolerance, Functional Reconstruction materially improves acceptance in 37 of 64 matched task cells, leaves 26 practically unchanged, and materially decreases one. We release our implementation and full evaluation artifact at \url{https://github.com/swyhahaha/FunctionalMLA}.
\end{abstract}

\section{Introduction}

Large language model inference is increasingly shaped by memory movement rather than arithmetic. During autoregressive decoding, every generated token reads a growing key-value (KV) cache, so serving latency and throughput depend heavily on how efficiently this cache is represented and accessed. Grouped-query attention (GQA) reduces the number of KV heads relative to multi-head attention, but it still stores explicit keys and values. Multi-head latent attention (MLA), used in DeepSeek-style models, goes further: it caches a low-dimensional latent vector and a small positional component, then computes attention through absorbed projections \citep{deepseek2024v2}. Recent conversion methods make this design practical without pretraining an MLA model from scratch: TransMLA converts existing GQA/MHA checkpoints, while MHA2MLA provides a related MHA-to-MLA migration path \citep{transmla2025,mha2mla2025}.

This paper studies a failure mode that is easy to miss if one only measures standalone decoding. A converted MLA model may be an efficient generator but still a poor \emph{draft model}. In speculative decoding, a draft proposes several tokens and a target model verifies them in parallel \citep{leviathan2023fast,chen2023accelerating}. When draft proposals agree with target verification, many tokens are accepted and the target advances by multiple positions per verification step. When conversion changes proposal probabilities or rankings, verification rejects more tokens and the draft becomes overhead.

In this paper, we introduce a new functional reconstruction method to improve draft acceptance and output-token speed during speculative decoding. It is a converter-agnostic, training-time-only post-conversion stage, requiring no verifier logits, verifier activations, benchmark labels, or generated targets, and making no change to the converted cache structure or inference graph. Starting from a TransMLA- or MHA2MLA-converted checkpoint, it trains the converter-introduced query and KV projections so that the complete MLA attention modules reproduce the post-$W_O$ responses of the frozen original MHA/GQA blocks on calibration hidden states. Across the 64 matched task cells in Table~\ref{tab:full-matrix}, our method yields 37 material acceptance improvements, 26 practically unchanged outcomes, and one material decrease; 12 of the 37 improved cells also show a material throughput increase. These results span TransMLA and MHA2MLA, HF and vLLM, and the Llama and Qwen model families.

Our contributions are threefold:

\noindent\textbf{(1) Conversion and drafting are distinct objectives.}
Structural conversion can succeed while the proposal agreement needed for draft acceptance is lost.

\noindent\textbf{(2) Draft quality can be optimized entirely at training time.}
Our functional objective uses calibration states and the frozen original attention block, without verifier supervision or inference-time changes.

\noindent\textbf{(3) Training-time functional reconstruction is converter-agnostic.}
The same objective improves or remains within the reporting tolerance in 63 of
64 matched task cells spanning TransMLA/MHA2MLA, Llama/Qwen, and HF/vLLM.

\section{Related Work}

\textbf{KV-cache efficient attention.}
Autoregressive Transformers retain a key and value state for each prior token and layer during decoding \citep{vaswani2017attention}. MQA and GQA reduce this state by sharing KV heads across query heads \citep{shazeer2019fast,ainslie2023gqa}. Other approaches preserve the original attention representation but reduce its runtime footprint: PagedAttention virtualizes explicit KV blocks \citep{kwon2023vllm}, H$_2$O and StreamingLLM retain selected cache entries for long sequences \citep{zhang2023h2o,xiao2023streamingllm}, and KIVI quantizes the cache \citep{liu2024kivi}. FlashAttention and FlashAttention-2 instead optimize exact attention kernels without changing the cached representation \citep{dao2022flashattention,dao2023flashattention2}. MLA is architecturally distinct: it replaces the explicit KV cache with a learned latent cache, so it can reduce the representation itself rather than only its storage or execution cost \citep{deepseek2024v2,deepseek2024v3}.

\textbf{MLA and conversion.}
MLA was introduced in DeepSeek-V2 and retained in DeepSeek-V3 as an inference-oriented attention architecture \citep{deepseek2024v2,deepseek2024v3}. Its migration from pretrained MHA/GQA is nontrivial because both low-rank KV compression and rotary position handling can perturb the original attention map \citep{su2024roformer}. TransMLA uses RoPE decomposition with low-rank QKV conversion \citep{transmla2025}, whereas MHA2MLA combines partial-RoPE removal with joint low-rank approximation \citep{mha2mla2025}. These methods solve the initialization problem with converter-local reconstruction objectives. Our E2E procedure is complementary: it begins from either converted checkpoint and uses original MHA/GQA attention outputs on calibration states as the functional target that matters for a speculative draft.

\textbf{Speculative decoding.}
Speculative decoding uses a cheap draft to propose multiple tokens and an exact target-side correction to preserve the target distribution \citep{leviathan2023fast,chen2023accelerating}. Tree-based methods enlarge the parallel candidate set through fixed or hardware-aware draft trees \citep{miao2023specinfer,chen2024sequoia}; Medusa uses auxiliary decoding heads \citep{cai2024medusa}; and EAGLE/EAGLE-2 draft in feature space with static or context-adaptive trees \citep{li2024eagle,li2024eagle2}. Self-speculative approaches reuse the target model through layer skipping or early exit \citep{zhang2023draftverify,elhoushi2024layerskip}, while Lookahead decoding uses Jacobi-style parallel updates \citep{fu2024lookahead}. These works alter proposal generation or verification scheduling. We instead improve the proposal fidelity of an MLA-converted draft, which is complementary to those designs and directly raises the acceptance on which their speedups depend.

\section{Preliminary: Why Converted MLA Is Not Draft-Preserving}

\subsection{MHA/GQA Attention}
Let $h_t \in \mathbb{R}^{D}$ be the hidden state at position $t$, where $D$
denotes the model hidden dimension, with $H$ query heads and $G$ KV groups.
We use GQA notation below; standard MHA is the special case $G=H$ with one
KV group per query head. For query head $i$ in group $g(i)$, GQA computes
\begin{align}
q_{t,i} &= W^Q_i h_t, &
k_{j,g(i)} &= W^K_{g(i)} h_j, &
v_{j,g(i)} &= W^V_{g(i)} h_j .
\end{align}
Let $R_t$ be the RoPE rotation at position $t$ and define the relative
rotation $R_{tj}=R_t^\top R_j$ \citep{su2024roformer}. With key dimension
$d_k$, scaled causal attention is
\begin{align}
s^{\mathrm{GQA}}_{tij} &=
\frac{q_{t,i}^{\top}R_{tj}k_{j,g(i)}}{\sqrt{d_k}}, \\
a^{\mathrm{GQA}}_{tij} &=
\frac{\exp(s^{\mathrm{GQA}}_{tij})}
{\sum_{u\leq t}\exp(s^{\mathrm{GQA}}_{tiu})}, \\
o^{\mathrm{GQA}}_{t,i} &= \sum_{j \leq t} a^{\mathrm{GQA}}_{tij} v_{j,g(i)} .
\end{align}

\subsection{Latent KV Bottleneck}
An MLA layer stores a latent cache $c_j=W^{DKV}h_j \in \mathbb{R}^{r}$ and a
small positional key $k^{PE}_j$. The non-positional keys and values of all KV
groups are stacked and produced from the latent:
\begin{equation}
\begin{bmatrix}
\hat{k}^{nope}_{j} \\
\hat{v}_{j}
\end{bmatrix}
=
W^{UKV} W^{DKV} h_j .
\end{equation}
Here $W^{DKV}\in\mathbb{R}^{r\times D}$ and
$W^{UKV}\in\mathbb{R}^{m\times r}$, where
$m=G(d_k^{nope}+d_v)$. Exact reproduction of this stacked linear map for all
hidden states requires
\begin{equation}
\begin{bmatrix}
W^{K,nope} \\
W^V
\end{bmatrix}
=
W^{UKV} W^{DKV}.
\label{eq:rank-condition}
\end{equation}
The right-hand side has rank at most $r$. Therefore exact equivalence requires
\begin{equation}
\mathrm{rank}\!\left(
\begin{bmatrix}
W^{K,nope} \\
W^V
\end{bmatrix}
\right) \leq r .
\label{eq:rank-bound}
\end{equation}
Whenever this rank exceeds $r$, the chosen compressed factorization cannot
reproduce the stacked K/V map exactly and must leave a residual. This
conditional statement concerns direct map reconstruction at the selected
rank: it neither rules out every functionally equivalent attention
parameterization nor contradicts exact MLA rewrites at a sufficiently large
rank. It shows why an aggressively compressed, cache-valid conversion is
generally an approximation of the original K/V projections.

\begin{figure*}[t]
\centering
\begin{tikzpicture}
\node[anchor=south west,inner sep=0] (overview) at (0,0) {
  \includegraphics[width=\textwidth]{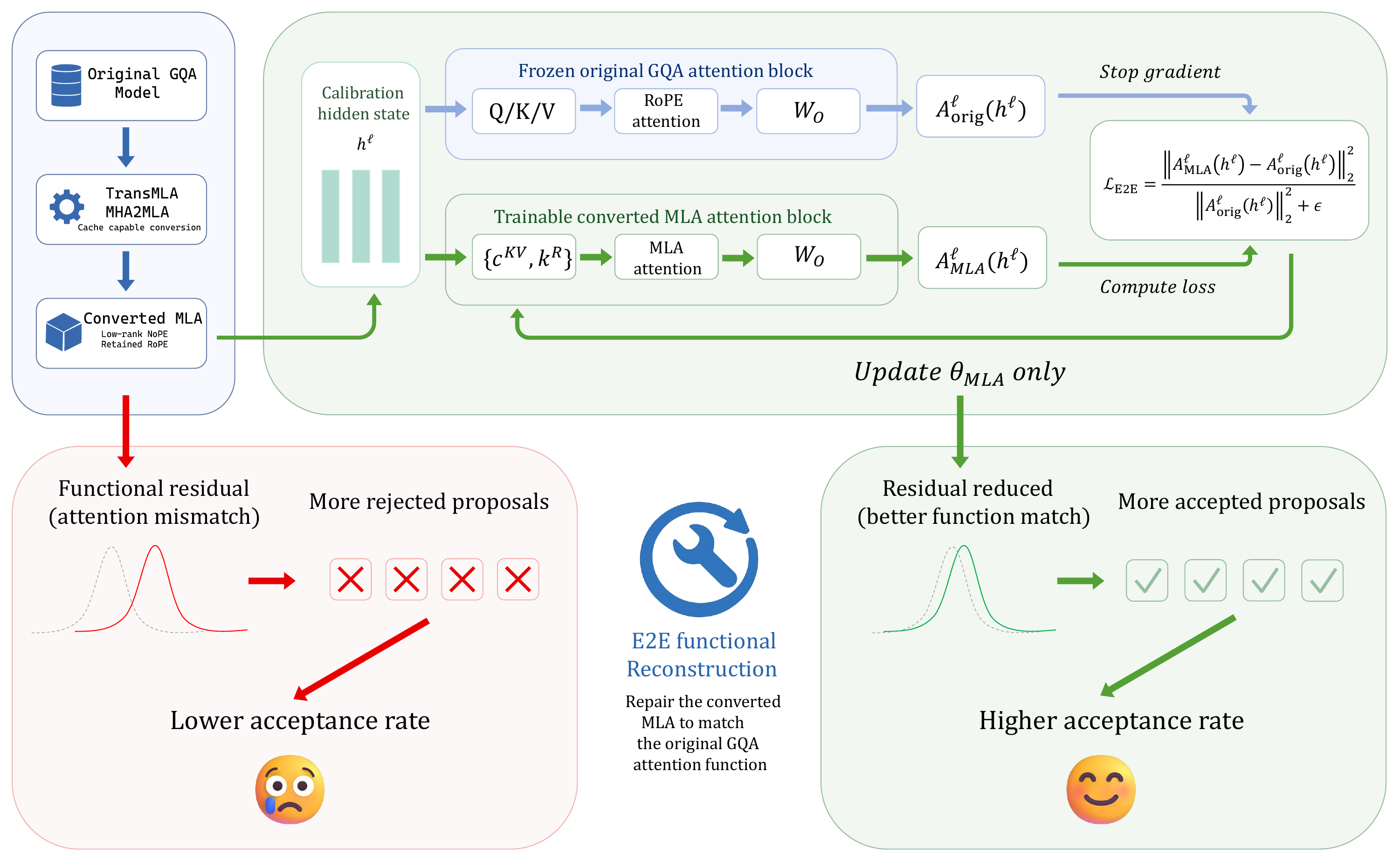}
};
\begin{scope}[x={(overview.south east)},y={(overview.north west)}]
\node[
  fill=white,
  minimum width=0.182\textwidth,
  minimum height=0.074\textwidth,
  inner sep=1pt
] at (0.878,0.792) {
  \resizebox{0.174\textwidth}{!}{$
  \mathcal{L}_{\mathrm{E2E}}^{(\ell)}
  =
  \dfrac{
    \left\|M\odot
    \left(A^\ell_{\mathrm{MLA}}-\operatorname{sg}[A^\ell_{\mathrm{orig}}]\right)
    \right\|_F^2
  }{
    D\sum_t M_t
  }
  $}
};
\end{scope}
\end{tikzpicture}
\caption{Converter-agnostic E2E functional reconstruction.
A TransMLA or MHA2MLA conversion first supplies a cache-capable MLA initialization.
For each calibration hidden state $h^\ell$, the frozen original GQA block and the trainable converted MLA block produce post-$W_O$ outputs.
E2E minimizes their masked mean-squared error, stops gradients through the original path, and updates only the converter-introduced query and KV projections.
The lower schematic illustrates the empirical pathway tested: reducing the residual can preserve proposal rankings and improve acceptance.}
\label{fig:e2e-overview}
\end{figure*}

\subsection{RoPE Handling and Acceptance}
The rank bottleneck is not the only mismatch. Conversion pipelines must also
choose how to handle RoPE. To make the resulting error explicit, split an
original key as $k_j=k_j^{keep}+k_j^{drop}$, where $k_j^{keep}$ remains on the
rotary path and $k_j^{drop}$ is moved to a non-positional path. Let the
converted query be $\hat q_t=q_t+\delta q_t$ and the reconstructed retained key
be $\hat k_j^{keep}=k_j^{keep}+\delta k_j$. Under this representative routing,
the original logit uses $q_t^\top R_{tj}(k_j^{keep}+k_j^{drop})/\sqrt{d_k}$,
whereas the converted logit uses
$(q_t+\delta q_t)^\top[R_{tj}(k_j^{keep}+\delta k_j)+k_j^{drop}]/\sqrt{d_k}$.
Their exact difference is
\begin{equation}
\begin{aligned}
\hat s_{tj}-s_{tj}
=\frac{1}{\sqrt{d_k}}\Big[
&q_t^\top R_{tj}\delta k_j
+q_t^\top(I-R_{tj})k_j^{drop}\\
&+\delta q_t^\top R_{tj}(k_j^{keep}+\delta k_j)
+\delta q_t^\top k_j^{drop}
\Big].
\end{aligned}
\label{eq:logit-error}
\end{equation}
This decomposition exposes retained-key reconstruction error, positional-path
error, and first-order query perturbation; the third term also contains their
interaction. Softmax maps the logit residual into attention-weight
error, and the output residual additionally inherits value reconstruction
error.

Speculative decoding makes this residual operationally visible. At a fixed
context $x$, write $p_D^x(y)=p_D(y\mid x)$ and similarly for $p_T^x$.
Stochastic rejection-sampling verification has one-token expected acceptance
\citep{leviathan2023fast,chen2023accelerating}
\begin{equation}
\begin{aligned}
\alpha_{\mathrm{sample}}(x)
&=\sum_y \min\!\left(p_D^x(y),p_T^x(y)\right)\\
&=1-\mathrm{TV}\!\left(p_D^x,p_T^x\right).
\end{aligned}
\label{eq:acceptance-overlap}
\end{equation}
Our experiments instead use greedy decoding, whose one-token acceptance event
is
\begin{equation}
\alpha_{\mathrm{greedy}}(x)
=
\mathbf{1}\!\left[
\arg\max_y p_D^x(y)=\arg\max_y p_T^x(y)
\right].
\label{eq:greedy-acceptance}
\end{equation}
MLA conversion can therefore hurt stochastic overlap or flip the top-ranked
token under greedy verification. Neither quantity is guaranteed to vary
monotonically with a local attention residual; functional reconstruction is a
training-time proxy intended to preserve the original draft's conditional
behavior, whose acceptance effect is measured empirically below.

\section{Method}

Figure~\ref{fig:e2e-overview} separates conversion, reconstruction, and
downstream acceptance. TransMLA or MHA2MLA first supplies a cache-capable
low-rank MLA initialization with its converter-defined retained-RoPE path.
During reconstruction, the same layer input is sent through a frozen original
GQA attention block and the corresponding converted MLA block. Their
post-$W_O$ responses define a masked mean-squared functional error. Gradients
are stopped through the original path and applied only to the
converter-introduced query and KV projections; $W_O$ remains frozen. Direct
conversion can leave a residual that changes proposal rankings, whereas E2E
reduces this residual without changing the cache topology. Its effect on
acceptance is empirical rather than guaranteed by the local loss. The target
verifier is not part of this optimization. Partial RoPE reconstruction remains
the matched converter-local baseline below, not a second branch of the E2E
training graph.

\subsection{Functional Reconstruction}
For layer $\ell$ and calibration minibatch $b$, let
$H_b^\ell\in\mathbb{R}^{T_b\times D}$ contain the cached layer inputs and let
$M_b\in\{0,1\}^{T_b}$ mark valid tokens, broadcast over the hidden dimension.
The implementation optimizes each layer independently with
\begin{equation}
\mathcal{L}_{\mathrm{E2E}}^{(\ell)}
=
\mathbb{E}_{b}
\left[
\frac{
\left\|M_b\odot\left(
A^\ell_{\mathrm{MLA}}(H_b^\ell)
-\operatorname{sg}\!\left[A^\ell_{\mathrm{orig}}(H_b^\ell)\right]
\right)\right\|_F^2
}{
D\sum_t M_{b,t}
}
\right].
\label{eq:e2e-loss}
\end{equation}
Thus, the denominator $D\sum_t M_{b,t}$ averages the squared error over every
hidden channel of every valid token. Here $\operatorname{sg}$ denotes
stop-gradient, and $A^\ell$ is the output of
attention layer $\ell$ after its fixed output projection. Unlike a
converter-local projection reconstruction objective, this masked MSE asks
whether the complete converted attention block behaves like the original
MHA/GQA attention on real calibration hidden states. In all reported Llama and
Qwen experiments, $A_{\mathrm{orig}}=A_{\mathrm{GQA}}$; the MHA case is
covered by $G=H$. The loss does not require the rank condition in
Eq.~\eqref{eq:rank-bound} to hold exactly; it searches within the fixed
compressed MLA family for a closer functional match. The term E2E refers to
the full attention function from layer input to post-output-projection output,
so the objective jointly covers RoPE-handling and low-rank KV conversion
errors.

\subsection{Converter-Agnostic Initialization}
Let $\mathcal{C}$ denote a base MHA/GQA-to-MLA converter. E2E takes $\mathcal{C}(\theta_{\mathrm{orig}})$ as an initialization but does not reuse $\mathcal{C}$'s local conversion objective. It requires only the original attention block, the initialized MLA block, and calibration hidden states. Consequently, the same E2E objective can refine TransMLA and MHA2MLA checkpoints while preserving the cache topology produced by the selected converter. Our evaluation instantiates both choices of $\mathcal{C}$ and compares Functional Reconstruction with the matched Partial RoPE reconstruction for every model, backend, and task combination.

\section{Experiments}

\subsection{Experimental Setup}

\paragraph{Models and conversion methods.}
We study two model families and two draft sizes per family. Llama-1B and
Llama-3B draft for Llama-8B, while Qwen-1.5B and Qwen-3B draft for Qwen-7B.
For each pair, we construct MLA drafts with both TransMLA and MHA2MLA. This
design tests whether the proposed objective transfers across model families,
compression pipelines, and draft capacities rather than depending on one
converter-specific parameterization.

\paragraph{Comparative methods.}
We report three draft forms. \emph{Original GQA} is the unconverted draft and
serves as a reference for the acceptance available before introducing an MLA
bottleneck. \emph{Partial RoPE reconstruction} is the matched converter-local
baseline. \emph{Functional Reconstruction (ours)} begins from the same
converted initialization and retains exactly the same attention topology,
latent-cache budget, and inference interface, but optimizes
Eq.~\eqref{eq:e2e-loss}. Consequently, the Partial RoPE--Functional comparison
isolates the reconstruction objective under a fixed model, converter, backend,
and task. Partial and Functional Reconstruction update the same
converter-introduced query and KV projections; output projections and
non-attention parameters remain frozen.

\subsection{Evaluation Protocol}

\paragraph{Benchmarks and decoding.}
We evaluate HumanEval~\cite{chen2021evaluating}, Alpaca~\cite{taori2023alpaca,wang2023selfinstruct},
Natural Questions (NQ)~\cite{kwiatkowski2019natural}, and CNN/DailyMail
(CNN/DM)~\cite{hermann2015teaching,nallapati2016abstractive}, representing code generation, instruction following, open-domain
question answering, and summarization. Each benchmark uses a fixed 200-prompt
manifest and a maximum of 128 generated tokens. HumanEval contains its 164
canonical problems plus 36 deterministic repeats to match the common sample
count; the other manifests contain 200 unique prompts. All runs use BF16,
batch size one, greedy decoding, and seed 42. The proposal length is
$\gamma=4$ for the 1B/1.5B drafts and $\gamma=3$ for the 3B drafts.

\paragraph{Systems and metrics.}
We evaluate Hugging Face (HF) and vLLM implementations using token acceptance
and generated-output tok/s. Partial and Functional Reconstruction use identical
runtime configurations. vLLM uses the supported converter-specific tensor
parallelism (TP2 for TransMLA and TP1 for MHA2MLA), so absolute throughput is
not compared across converters. The benchmark contains 192 method--task
configurations and 38,400 generations, with 200 valid outputs per
configuration. For reporting, differences within 0.5 percentage points (pp) in
acceptance or 0.5 tok/s are treated as practically unchanged; these are
descriptive tolerances, not confidence intervals.

\section{Results and Analysis}

\begin{table*}[!t]
\centering
\begingroup
\small
\setlength{\tabcolsep}{0.85mm}
\begin{tabular}{@{}lllcccc@{\hspace{2.0mm}}cccc@{}}
\toprule
& & &
\multicolumn{4}{c}{\textbf{(a) Llama-1B $\rightarrow$ Llama-8B}} &
\multicolumn{4}{c}{\textbf{(c) Qwen-1.5B $\rightarrow$ Qwen-7B}} \\
\cmidrule(lr){4-7}\cmidrule(lr){8-11}
Converter & Backend & Draft & HE & Alp. & NQ & CNN/DM & HE & Alp. & NQ & CNN/DM \\
\midrule
TransMLA & HF & Original
  & 77.75/45.10 & 62.38/39.84 & 59.97/38.28 & 53.46/33.77
  & 88.29/34.92 & 56.50/26.15 & 44.99/22.46 & 47.89/22.56 \\
& & Partial
  & 67.61/37.73 & 54.90/33.86 & 47.96/31.09 & 39.87/26.11
  & 81.78/30.48 & 51.00/22.26 & 38.97/18.76 & 41.71/19.67 \\
& & \textbf{Ours}
  & \textbf{71.53}/\textbf{40.50} & \textbf{57.33}/\textbf{34.43} & \textbf{51.74}/\textbf{33.10} & \textbf{42.08}/\textbf{27.23}
  & 81.90/30.62 & 51.46/22.08 & 39.20/19.01 & 42.01/19.80 \\
& vLLM & Original
  & 87.64/80.93 & 79.64/77.89 & 76.50/77.51 & 74.53/63.39
  & 90.04/63.39 & 77.18/63.15 & 68.55/60.56 & 68.81/54.55 \\
& & Partial
  & 55.10/50.47 & 58.37/52.53 & 54.36/52.44 & 43.54/42.90
  & 15.75/27.50 & 16.23/29.10 & 17.67/28.35 & 12.88/25.63 \\
& & \textbf{Ours}
  & \textbf{57.18}/50.49 & \textbf{62.09}/\textbf{53.53} & \textbf{57.40}/\textbf{53.20} & \textbf{47.77}/\textbf{44.57}
  & 15.94/27.50 & \textbf{17.23}/29.24 & 17.48/28.48 & 12.73/25.45 \\
MHA2MLA & HF & Original
  & 77.75/45.10 & 62.38/39.84 & 59.97/38.28 & 53.46/33.77
  & 88.29/34.92 & 56.50/26.15 & 44.99/22.46 & 47.89/22.56 \\
& & Partial
  & 76.20/34.52 & 61.50/30.25 & 57.98/28.94 & 51.30/25.13
  & 81.95/24.00 & 49.84/17.27 & 38.55/14.55 & 43.39/15.45 \\
& & \textbf{Ours}
  & 76.20/34.17 & 61.50/29.92 & 57.98/28.29 & 51.30/24.97
  & \textbf{83.93}/\textbf{24.78} & \textbf{51.43}/17.59 & \textbf{39.55}/14.70 & \textbf{44.33}/15.35 \\
& vLLM & Original
  & 87.07/83.84 & 79.42/75.08 & 76.35/73.98 & 74.40/65.70
  & 90.01/67.87 & 76.89/61.94 & 68.21/56.97 & 68.79/52.20 \\
& & Partial
  & 85.54/40.45 & 77.94/35.03 & 75.37/34.06 & 72.22/30.57
  & 88.43/27.51 & 74.06/20.08 & 64.80/18.05 & 66.36/17.97 \\
& & \textbf{Ours}
  & 85.54/40.45 & 77.94/35.23 & 75.37/34.06 & 72.22/30.96
  & 88.65/27.77 & \textbf{75.16}/20.52 & \textbf{65.58}/18.16 & \textbf{67.21}/\textbf{18.49} \\
\midrule
& & &
\multicolumn{4}{c}{\textbf{(b) Llama-3B $\rightarrow$ Llama-8B}} &
\multicolumn{4}{c}{\textbf{(d) Qwen-3B $\rightarrow$ Qwen-7B}} \\
\cmidrule(lr){4-7}\cmidrule(lr){8-11}
Converter & Backend & Draft & HE & Alp. & NQ & CNN/DM & HE & Alp. & NQ & CNN/DM \\
\midrule
TransMLA & HF & Original
  & 84.41/32.69 & 74.96/31.12 & 72.82/30.54 & 68.28/27.74
  & 89.83/29.14 & 64.71/23.70 & 54.76/21.37 & 58.82/21.59 \\
& & Partial
  & 72.30/26.51 & 62.42/23.93 & 51.07/21.49 & 49.64/20.10
  & 86.97/25.41 & 58.99/19.83 & 47.58/17.81 & 52.13/18.31 \\
& & \textbf{Ours}
  & \textbf{73.27}/26.87 & \textbf{63.12}/24.41 & \textbf{52.68}/21.94 & \textbf{50.32}/20.23
  & \textbf{87.92}/25.90 & \textbf{59.97}/19.94 & \textbf{48.27}/17.87 & \textbf{53.25}/18.02 \\
& vLLM & Original
  & 89.47/56.61 & 85.26/56.82 & 83.12/54.51 & 80.95/49.24
  & 91.53/49.59 & 78.94/46.96 & 71.03/50.86 & 71.66/42.10 \\
& & Partial
  & 56.18/36.26 & 66.50/36.74 & 56.63/36.58 & 47.75/33.39
  & 25.30/25.30 & 6.83/26.64 & 10.95/26.05 & 17.58/23.67 \\
& & \textbf{Ours}
  & \textbf{59.40}/36.46 & \textbf{68.22}/\textbf{38.27} & \textbf{59.01}/\textbf{37.96} & \textbf{49.13}/33.75
  & 25.35/25.20 & 6.73/26.36 & \textbf{12.73}/26.06 & 17.03/23.60 \\
MHA2MLA & HF & Original
  & 84.41/32.69 & 74.96/31.12 & 72.82/30.54 & 68.28/27.74
  & 89.83/29.14 & 64.71/23.70 & 54.76/21.37 & 58.82/21.59 \\
& & Partial
  & 80.51/23.85 & 68.65/21.66 & 64.67/20.53 & 59.11/18.89
  & 87.41/21.13 & 60.36/16.41 & 51.22/15.19 & 54.00/15.07 \\
& & \textbf{Ours}
  & 80.51/23.73 & 68.65/21.70 & 64.67/20.80 & 59.11/19.16
  & \textbf{89.26}/21.51 & \textbf{61.98}/16.64 & \textbf{52.52}/15.10 & \textbf{56.29}/15.39 \\
& vLLM & Original
  & 88.99/60.79 & 85.10/58.71 & 82.93/57.14 & 80.85/49.88
  & 91.79/55.54 & 78.63/51.45 & 70.85/46.73 & 71.78/44.68 \\
& & Partial
  & 84.96/27.96 & 81.25/26.17 & 77.68/25.26 & 73.09/22.63
  & 90.21/24.32 & 75.59/19.24 & 67.39/16.21 & 68.92/17.20 \\
& & \textbf{Ours}
  & 84.96/28.22 & 81.25/26.20 & 77.68/24.86 & 73.09/22.44
  & \textbf{90.96}/23.31 & \textbf{77.59}/19.41 & \textbf{68.59}/\textbf{17.69} & \textbf{70.36}/17.45 \\
\bottomrule
\end{tabular}

\endgroup
\caption{Complete results over all 192 method--task configurations (200 prompts
per task). Cells report acceptance (\%) / output tok/s. Original, Partial, and
Ours denote unconverted GQA, Partial RoPE, and Functional Reconstruction,
respectively. Bold marks Ours only when it exceeds matched Partial by more than
0.5 pp in acceptance or 0.5 tok/s in throughput; smaller differences are left
unbolded. Comparisons are within converter--backend blocks.}
\label{tab:full-matrix}
\end{table*}

\subsection{Matched-Cell Effectiveness}

Table~\ref{tab:full-matrix} reports detailed results for every evaluated
configuration. Partial RoPE and Functional Reconstruction are compared within
each fixed model--converter--backend--task cell. Under the reporting tolerance,
37 of 64 cells show a material acceptance improvement, 26 are practically
unchanged, and one materially decreases. Thus, 37 of the 38 changes outside the
tolerance favor Functional Reconstruction.

The sole material decrease is $-0.55$ pp on CNN/DM for Qwen-3B with TransMLA
under vLLM. The largest improvements are $+4.23$ pp on CNN/DM
(Llama-1B, TransMLA, vLLM), $+3.92$ pp on HumanEval and $+3.78$ pp on NQ
(Llama-1B, TransMLA, HF), and $+3.72$ pp on Alpaca
(Llama-1B, TransMLA, vLLM). Every Partial--Ours pair holds latent rank,
retained-RoPE width, cache budget, proposal length, and inference graph fixed.
The material differences therefore isolate the reconstruction objective within
the same compressed MLA parameterization rather than an increase in draft
capacity.

\subsection{Cross-Factor Analysis}

\paragraph{Task breadth reflects a shared functional correction.}
All four benchmarks contain material gains, but the small differences in their
counts are better explained by model--converter interactions than by task
identity. HumanEval contains eight material improvements and eight practically
unchanged cells; Alpaca and NQ each contain 10 improvements and six unchanged
cells; CNN/DM contains nine improvements, six unchanged cells, and one
borderline decrease. Each task receives the same four improvements from
Llama--TransMLA and the same four unchanged comparisons from
Llama--MHA2MLA. The remaining variation comes entirely from the Qwen paths.
Thus, no single domain---code, instruction following, open-domain QA, or
summarization---accounts for the overall result. This task breadth is expected
from a reconstruction objective trained on calibration hidden states rather
than benchmark labels: it corrects a shared attention-block mapping, and the
benefit transfers when a downstream task visits the corrected state region.
The isolated CNN/DM deviation is only 0.05 pp beyond the descriptive tolerance
and does not constitute a repeated task-level regression.

\paragraph{Model family and converter exhibit a clear interaction.}
Functional Reconstruction is applicable to both converters, but its measurable
headroom is converter- and family-dependent. All 16 Llama--TransMLA cells
improve, by $+0.68$ to $+4.23$ pp, whereas all 16 Llama--MHA2MLA cells are
unchanged task by task. Qwen shows the complementary pattern:
Qwen--MHA2MLA improves in 15 of 16 cells, with the remaining cell within
tolerance, while Qwen--TransMLA improves in six, leaves nine unchanged, and
contains the borderline CNN/DM deviation. This cross-over rules out a
converter-specific explanation for the method: the same post-$W_O$ objective
helps TransMLA most consistently for Llama and MHA2MLA most consistently for
Qwen. It also rules out the stronger claim that conversion method is
irrelevant. Rather, converter initialization determines which functional
residual remains inside the trainable query and KV projections. An exact
Llama--MHA2MLA tie means that the refinement does not alter enough proposal
rankings to change acceptance under this protocol; the broad Qwen--MHA2MLA and
Llama--TransMLA gains indicate residual errors that the same training-only
objective can remove.

\paragraph{Backend dependence is localized rather than universal.}
Under HF, 20 cells materially improve and 12 remain within tolerance; under
vLLM, 17 improve, 14 remain within tolerance, and one is borderline. The
aggregate difference is concentrated in Qwen--TransMLA. For HF,
Qwen-1.5B--TransMLA is unchanged on all four tasks, whereas
Qwen-3B--TransMLA improves on all four. Under vLLM, only Alpaca improves for
Qwen-1.5B--TransMLA, and only NQ improves for Qwen-3B--TransMLA. In contrast,
the Llama pattern is invariant across backends---TransMLA improves and
MHA2MLA ties---and Qwen--MHA2MLA improves in all eight HF cells and seven of
eight vLLM cells. The objective therefore transfers across both runtime
implementations, but the realized operator can limit how much of the learned
correction survives deployment. This is especially relevant because vLLM uses
TP2 for TransMLA and TP1 for MHA2MLA; the absolute numbers are not a controlled
cross-converter comparison.

\paragraph{Draft size alone does not explain the gains.}
The 1B/1.5B partition contains 16 material improvements and 16 unchanged
cells, while the 3B partition contains 21 improvements, 10 unchanged cells,
and the borderline deviation. This apparent size difference is family-specific:
Llama-1B and Llama-3B each contribute exactly eight improvements and eight
ties, whereas Qwen-1.5B contributes eight improvements and eight unchanged
cells and Qwen-3B contributes 13 improvements, two unchanged cells, and the
borderline point. The additional 3B gains therefore come from Qwen's
interaction with the two converter initializations, not from a general scaling
law. Moreover, the smaller and 3B drafts use $\gamma=4$ and $\gamma=3$,
respectively, so their absolute acceptances are not directly comparable.
Within-cell Partial--Ours differences remain the controlled evidence.

\paragraph{Recovery relative to Original identifies the method boundary.}
The Original GQA rows diagnose conversion damage but are not formal upper
bounds. Qwen--MHA2MLA exhibits a moderate, recoverable gap: on CNN/DM with HF,
Qwen-3B changes from 58.82\% (Original) to 54.00\% (Partial) and 56.29\%
(Ours), and on Alpaca with vLLM it changes from 78.63\% to 75.59\% and
77.59\%. Llama--TransMLA also shows partial recovery from larger losses:
Llama-1B on HumanEval with HF changes from 77.75\% to 67.61\% and 71.53\%,
while its CNN/DM vLLM cell changes from 74.53\% to 43.54\% and 47.77\%.
However, initial damage and recoverability are not monotonic. For
Qwen-3B--TransMLA with vLLM, NQ changes from 71.03\% to 10.95\% and 12.73\%;
the gain is real but small relative to the conversion loss. Functional
matching can correct approximation error that remains representable by a valid
converted block, but it cannot create missing latent rank or repair an
incompatible tensor-parallel or kernel realization. This distinction explains
both the broad matched gains and the severe gaps that remain.

\subsection{Throughput and the Quality--Efficiency Trade-off}

Functional Reconstruction is training-only and introduces no additional
inference operation, parameter branch, or cache state. Under the throughput
reporting tolerance, 12 cells are materially faster, 50 are practically
unchanged, and two are materially slower; changes outside the tolerance range
from $-1.01$ to $+2.77$ tok/s. Among the 37 cells with a material acceptance
gain, 12 are also materially faster, 24 are unchanged in throughput, and one is
slower. Without repeated timing runs, the individual tok/s differences should
be treated as a combination of acceptance effects and run-level variation. The
controlled conclusion is that improved acceptance does not require an added
inference operation or a changed cache budget.

The present measurements do not establish an end-to-end speedup over the
unconverted draft. Absolute Original GQA--MLA throughput differences combine
model fidelity with GQA/MLA kernel maturity, converter-specific tensor
parallelism, and backend integration. Optimized MLA kernels and a common
runtime configuration remain necessary to translate the model-side gain into a
clean system-level speedup over native GQA.

\section{Discussion}

\paragraph{(1) Conversion and drafting define different success criteria.}
The results expose three nested notions of success. \emph{Structural validity}
asks whether an MHA/GQA checkpoint has been mapped to an executable MLA
parameterization with the intended latent rank, RoPE path, and cache budget.
\emph{Functional fidelity} asks whether that parameterization realizes the same
attention-block mapping on states encountered by the model. \emph{Draft
utility} asks whether the resulting token distribution agrees with a separate
verifier closely enough to produce long accepted runs. A converter can satisfy
the first criterion while failing the latter two: the severe absolute
acceptance losses in several executable TransMLA paths make this separation
explicit. Conversely, the matched Partial--Ours gains change neither structure
nor inference cost, demonstrating that structural conversion does not exhaust
the optimization problem.

This hierarchy also clarifies what should be preserved. Factor-level error in
$Q$, $K$, or $V$ is only an indirect proxy because attention composes low-rank
projections, RoPE, softmax, value aggregation, and $W_O$ before affecting the
residual stream. Small errors can cancel, while a localized logit perturbation
can cross a token-ranking margin and sharply reduce acceptance. Conversion
should therefore be evaluated not only by cache validity or standalone
generation, but also by an interface-level functional metric and matched draft
acceptance. The Original GQA draft is a useful semantic anchor, not a formal
acceptance upper bound: speculative utility is agreement with the target model,
not self-fidelity to the pre-conversion draft. This distinction is the central
reason that MLA conversion and MLA draft construction should be treated as
related but separate tasks.

\paragraph{(2) Functional reconstruction provides a training-time control
knob for draft quality.}
The broad gains across code, instruction following, QA, and summarization show
that draft usefulness can be improved offline without labels from any of those
tasks and without placing the verifier in the optimization graph.
Post-$W_O$ matching is important here because it trains against the observable
output of the complete attention transformation rather than against one
converter-specific factorization. Calibration hidden states supply the
operating distribution, and the frozen original block supplies a stable
functional reference. In this sense, Functional Reconstruction is not another
cache design; it searches for a better function inside the fixed function class
and memory budget established by conversion.

Acceptance nevertheless remains a discrete, end-to-end consequence of many
layers, whereas Eq.~\eqref{eq:e2e-loss} is a continuous, layer-local surrogate.
The borderline negative cell and the unrecovered gaps are therefore
informative: reducing local block error is generally useful, but need not move
the final token ranking monotonically. Later layers may attenuate a residual,
amplify it, or project it onto a direction with little decision relevance.
This suggests a principled path beyond uniform MSE, including
layer-sensitivity weighting, token-margin-aware reconstruction, or short
sequence-level functional matching, while retaining a verifier-free training
stage. Crucially, such refinement adds no inference branch or cache state.
It can improve the acceptance term in speculative speedup while leaving draft
cost fixed; whether that gain becomes wall-clock speedup remains a separate
kernel and runtime question.

\paragraph{(3) Generality comes from the functional interface.}
The Llama/Qwen cross-over in Table~\ref{tab:full-matrix} is particularly
revealing. The same objective is most consistently effective for
Llama--TransMLA and Qwen--MHA2MLA, while Llama--MHA2MLA has little measurable
headroom and Qwen--TransMLA is more backend-sensitive. Thus, generality comes
from placing the objective at a shared functional interface, not from assuming
that every converter leaves the same error. A converter fixes both the
representable MLA function class and its initialization within that class;
model architecture, rank allocation, and RoPE treatment determine the
remaining approximation residual. Functional Reconstruction can optimize the
portion of that residual reachable through the converted query and KV
projections, but it cannot recover information excluded by the latent rank.
Likewise, training cannot repair a backend that realizes different tensor
semantics from the calibrated module. The Qwen--TransMLA vLLM gap is therefore
best read as a precondition failure for reconstruction, not evidence that a
shared functional objective is tied to one converter.

\paragraph{Implications for MLA conversion pipelines.}
These observations motivate a three-level workflow. First, conversion should
establish a cache-efficient parameterization and verify RoPE, tensor-parallel,
and kernel correctness. Second, functional reconstruction should optimize the
converted attention map under that fixed deployment budget. Third, deployment
validation should measure matched acceptance and backend-specific throughput.
Keeping these levels separate makes failures diagnosable: structural tests
locate representation or runtime errors, functional tests expose recoverable
approximation error, and speculative evaluation measures decision-level value.
More broadly, latent rank should be viewed as a resource constraint rather
than a quality certificate. Future converters can use functional residuals to
allocate rank across layers, choose between initializations, and decide where
additional reconstruction capacity yields the largest draft-level return.

\section{Limitations}

Although the matrix contains 192 configurations, it remains modest in model
and data scale: two model families,
four draft--target pairs no larger than 8B, four 200-prompt benchmarks, a
128-token generation limit, and one fixed seed. Experiments with larger
targets, more architectures and converters, repeated seeds, and long-context
workloads are needed to establish how the conclusions scale beyond the present
setting. Moreover, Functional Reconstruction consistently refines Partial RoPE
conversion, but it usually recovers only part of the acceptance lost relative
to Original GQA and does not surpass the corresponding Original acceptance in
the evaluated matrix. The residual gap is especially large when the initial
converter--backend path is severely mismatched. Although Original GQA is not a
theoretical upper bound on target agreement, this empirical gap shows that the
current layer-local objective cannot fully overcome fixed-rank information loss
or runtime realization errors. Stronger sequence-level reconstruction and
conversion--runtime co-design are required for full recovery.

\section{Conclusion}

MLA conversion and draft usefulness are different objectives. A converted MLA
checkpoint may be a valid generator but a poor speculative draft because the
latent cache imposes rank constraints and RoPE handling perturbs attention
logits. E2E functional reconstruction targets this residual without changing
the MLA cache or using verifier supervision. Across 192 evaluated method--task
configurations, Functional Reconstruction materially improves acceptance in 37
of 64 matched task cells, leaves 26 practically unchanged, and materially
decreases one relative to Partial RoPE reconstruction. The result spans both
TransMLA and MHA2MLA, Llama and Qwen, and HF and vLLM, but the evaluation also
shows that reconstruction cannot rescue a severely mismatched converter/backend
path. Functional reconstruction is therefore a general post-conversion
refinement, while converter quality and optimized MLA kernels remain
prerequisites for strong speculative-decoding speedups.

{\small
\bibliography{reference}
}

\end{document}